\documentclass{article} % For LaTeX2e
\usepackage{iclr2025_conference,times}

\usepackage{amsmath,amsfonts,bm}

\def\eqref#1{equation~\ref{#1}}
\def\1{\bm{1}}

\def\vl{{\bm{l}}}

\def\vo{{\bm{o}}}

\def\vu{{\bm{u}}}

\def\vx{{\bm{x}}}
\def\vy{{\bm{y}}}

\def\mA{{\bm{A}}}
\def\mB{{\bm{B}}}
\def\mC{{\bm{C}}}
\def\mD{{\bm{D}}}

\def\mK{{\bm{K}}}

\DeclareMathAlphabet{\mathsfit}{\encodingdefault}{\sfdefault}{m}{sl}
\SetMathAlphabet{\mathsfit}{bold}{\encodingdefault}{\sfdefault}{bx}{n}

\def\sL{{\mathbb{L}}}

\def\sO{{\mathbb{O}}}

\newcommand{\R}{\mathbb{R}}

\usepackage[list=true]{subcaption}
\usepackage{wrapfig}
\usepackage{hyperref}
\usepackage{multirow}
\usepackage{multicol}
\usepackage{url}

\usepackage{tikz}
\usetikzlibrary{shapes.geometric, arrows, positioning, fit}
\usepackage{pgfplots}
\usepackage{pgfplotstable}
\pgfplotsset{compat=1.18}
\usepgfplotslibrary{groupplots}

\definecolor{darkred}{RGB}{165,0,38}
\definecolor{lightred}{RGB}{215,48,39}
\definecolor{darkorange}{RGB}{244,109,67}
\definecolor{lightorange}{RGB}{253,174,97}
\definecolor{lightblue}{RGB}{116,173,209}
\definecolor{mediumblue}{RGB}{69,117,180}
\definecolor{darkblue}{RGB}{49,54,149}
\definecolor{my-green}{RGB}{0,168,0}

\title{Efficiently Scanning and Resampling Spatio-Temporal Tasks with Irregular Observations}

\author{Bryce Ferenczi \& Michael Burke \\
Department of Electrical and Computer Systems Engineering\\
Monash University\\
Melbourne, Australia \\
\texttt{\{bryce.ferenczi,michael.g.burke\}@monash.edu} \\
\And
Tom Drummond \\
Computing and Information Systems \\
University of Melbourne \\
Melbourne, Australia \\
\texttt{\{tom.drummond\}@unimelb.edu}
}

\iclrfinalcopy % Uncomment for camera-ready version, but NOT for submission.
\begin{document}

\maketitle

\begin{abstract}

Various works have aimed at combining the inference efficiency of recurrent models and training parallelism of multi-head attention for sequence modeling. However, most of these works focus on tasks with fixed-dimension observation spaces, such as individual tokens in language modeling or pixels in image completion. To handle an observation space of varying size, we propose a novel algorithm that alternates between cross-attention between a 2D latent state and observation, and a discounted cumulative sum over the sequence dimension to efficiently accumulate historical information. We find this resampling cycle is critical for performance. To evaluate efficient sequence modeling in this domain, we introduce two multi-agent intention tasks: simulated agents chasing bouncing particles and micromanagement analysis in professional StarCraft II games. Our algorithm achieves comparable accuracy with a lower parameter count, faster training and inference compared to existing methods.

\end{abstract}

\section{Introduction}
% Setup the problem
Spatio-temporal modelling tasks with complex unstructured or semi-structured state and observation spaces can be identified in a variety of domains. Designing deep learning algorithms to excel at these tasks often requires deliberate handling of both the accumulation of knowledge from historical observations, and the summarization of current observations, which can be computationally expensive. This is a particular concern in domains such as motion prediction and behaviour modeling, which have real-time requirements and often need to be performed using relatively low compute resources available on an edge devices. Recurrency is a popular paradigm in deep learning as it naturally maps onto these problems, which are sequential and causal in nature. While iterative processing of data with recursion can be efficiently performed in $\mathcal{O}(1)$ with respect to the input sequence, transformer and convolution methods introduce compute and memory complexity correlations with the input sequence length of $\mathcal{O}(N^2)$ and $\mathcal{O}(N)$ respectively. However, transformer and convolution algorithms are more parallelizable at training time, efficiently utilizing hardware and invoking backward propagation paths that aren't correlated with the sequence length. This leads to a trade-off between compute efficiency at training or inference time.

% Setup current works
Proposals to address this trade-off can be grouped into various categories, each of which aim to take advantage of inference efficiency of recurrence and parallelization at training time. One avenue aims to address the $\mathcal{O}(N^2)$ space and compute complexity that multi-head attention invokes over the sequence by introducing variations with linearized or amortized attention calculations \cite{sun2023retnet,peng2023rwkv,katharopoulos2020linattn}. State space models (SSMs) \cite{mamba2,gu2022s4} are presented as a compelling alternative to transformers with demonstrated efficacy in a variety of long range dependency tasks and language modeling. SSMs can be formulated as a convolution for training time paralellism or as a recurrent model for efficient inference over long sequences (\cite{gu2022s4}). \cite{fu2023simple} propose to directly learn an unrolled state space model with an efficient long-convolution framework. In each case, these methods are evaluated on sequences with a fixed dimensional observation space $\mathcal{O}\in\R^d$, such as tokenized text, image pixels or audio spectrogram data. Focusing on these sequences may leave a blind-spot in tasks with an irregular observation spaces $\mathcal{O}\in\R^{d(t)}$ such those in as multi-agent interactions, where the number of agents and links between these may vary over time. %Zero-padding is a common approach to handle this type of complexity, but adds unnecessary computational burden.

% Setup our work
In this work, we investigate a range of encoding and state space modelling approaches for these settings, and propose a novel and efficient algorithm that utilizes a 2D latent state and alternates between input sampling, and accumulating historical information as a weighted sum (inclusive scan). This weighted sum can be performed efficiently in parallel on a GPU with an inclusive-scan \cite{merrill2016scan} during training, and incrementally during inference. We show that the resampling cycle is more effective than a continued self-attention block, or not alternating between accumulation and processing. This method natively supports a two dimensional latent state. We evaluate our method against a transformer encoder \citep{AttnIsAllYouNeed}, several RNN models \citep{chung2014gru,hochreiter1997lstm}, and Mamba2 \citep{mamba2}.

To test the efficacy of these training and inference efficient algorithms on tasks with more complex observation spaces, we use two multi-agent interaction benchmarks. The first is a ``gymnasium'' style simulation that involves agents chasing randomly assigned particles\footnote{Environment and path-planning algorithm derived from \href{https://www.doc.ic.ac.uk/~ajd/Robotics/RoboticsResources/planningmultirobot.py}{here}.}. The second benchmark is based on \textit{StarCraft II} (SC2), a real-time strategy video game, where we extract instances where the players are in combat. Each of these tasks involves a multidimensional time-varying observation space, where the model must analyse the motion of agents to identify their intended goal. 
% Contrib summary
In summary our contributions are as follows:
\begin{itemize}
    \item The introduction of two multi-agent interaction challenges to better evaluate sequence modeling algorithms with irregular observation spaces.
    \item A novel algorithm to efficiently address sequence modelling tasks with irregular observation spaces. We find that this algorithm achieves comparable accuracy to alternatives with a lower parameter count and improved throughput in training and inference regimes.
    \item The finding that resampling the observation space conditioned on accumulated sequence data is particularly beneficial in sequence modelling tasks.% future works on sequence modeling should consider this proposed accumulation and resampling cycle. 
\end{itemize}

% Heavy handed approaches such as using a transformer encoder across the observation and time dimensions becomes unreasonably expensive due to the quadratic complexity of the standard multi-head attention algorithm. This can be reduced by decoupling the observation and time dimensions by compressing the observation before accumulating over time. However, the time dimension is still susceptible to quadratic complexity. Recursive algorithms avoid this by incrementally updating a latent representation of the scene. This has previously been the typical paradigm of the family of recursive neural networks, RNN, LSTM and GRU. Furthermore, use the latent state the sampling algorithm on the current state of the scene.

\section{Background}

\textbf{Efficient and parallelisable sequence modeling} is of great interest to the research community, as attaining efficient utilisation of parallel computation at training time while performing inference efficiently are desirable attributes of sequence modeling algorithms. Previously, a trade-off had to be considered between using a recurrent neural network and a window-based algorithm such as a transformer or convolutional neural network. Recurrent methods utilize a hidden state to retain information between steps in the sequence, enabling $\mathcal{O}(1)$ compute and memory complexity with respect to the length of the sequence when performing incremental inference. However, incremental processing results in under-utilization of compute hardware and long back-propagation chains dependent on the sequence length, slowing the training process. Alternatively, transformer or convolution based methods are highly parallelisable, taking advantage of modern hardware, and the back-propagation chain is related to the depth of the model, rather than the length of the sequence. This results in greater training throughput. This comes at the cost of lack of incremental processing for inference. At inference time, a window of data is processed causing runtime compute and memory complexity of $\mathcal{O}(L)$ and $\mathcal{O}(L^2)$ for convolution and transformer methods respectively. 

\textbf{Spatio-temporal tasks} can be modeled as a state space $\mathcal{S}$ with an observation space $\mathcal{O}$ that evolves over time according to dynamics model $\mathcal{D}$. In some domains, $\mathcal{O}$ is a variably-sized set, $\sO_t$, that changes between environment instances, or over the duration of the sequence. This creates challenges in both concisely and effectively compressing a sequence of variably sized observations into some latent representation of the state of the system $\mathcal{S}_t':=\sO_{0\dots T}$. This can be exacerbated if the temporal duration of this task is indefinite as $T\in\mathbb{N}^+$.

As a particularly pertinent example of this class of problem, StarCraft II (SC2) is a real-time strategy game where players build an economy and military in order to defeat the opposing team. At a macro scale, SC2 follows a typical rock-paper-scissors approach where there are counter-strategies that can be employed against a given player. However, micromanagement in SC2 games is also a contributing factor to the performance of a player, with novices performing a few dozen actions per minute, whereas the typical professional completes hundreds of actions per minute. These actions are typically taken on variably sized sets of units, buildings, targets or objectives. Analysing the micromanagement of a player can potentially give insights into strategy used.

\section{Environments} \label{sec:env}

We describe the environments that are used to evaluate the efficacy of the algorithms first, as this will provide context for the design motivations in Section \ref{sec:method}. Each consist of two sets of agents which interact with one another with some objective. We use these environments to test whether models are capable of capturing relevant information from observations of a complex environment, in order to infer some properties from it. The predominant task is a target assignment problem, where an agent from one set is (possibly) assigned to target another. This task is similar to other intention recognition tasks which are pertinent in domains such as human robot interaction \citep{felip2022intuitive,ahmad2016bayesian}, that aim to infer and predict the objective of agents though observation. Each environment has an additional task to demonstrate generalisation of the architecture to multiple problems problems, beyond the target assignment problem.

\begin{figure}
    \centering
    \subcaptionbox{Chasing-Targets gymnasium environment. Robots (marked with a trajectory trail) are randomly allocated targets (\textcolor{blue}{blue} dots) to chase. When targets are reached (\textcolor{green}{Green} dots), robots are randomly assigned new targets.
    \label{fig:gym-img}}
    {\includegraphics[width=0.5\linewidth]{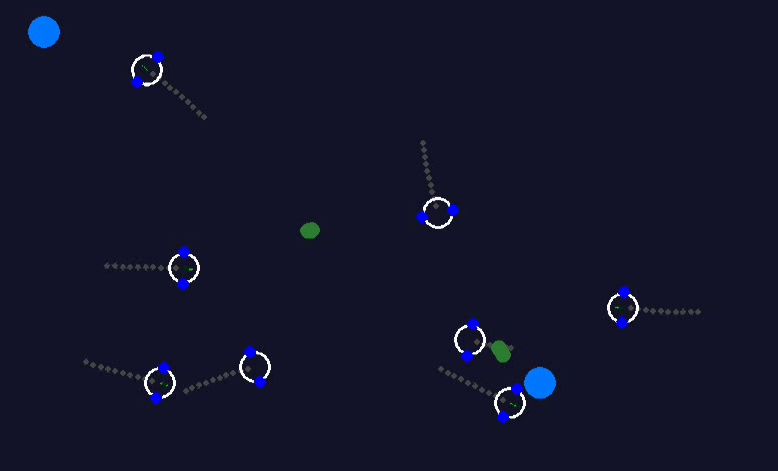}}
    \hfill
    \subcaptionbox{StarCraft II observation data. \textcolor{blue}{Blue} circles are player units, \textcolor{red}{red} circles are enemy units, \textcolor{green}{green} arrows are unit-target assignments and the \textcolor{gray}{grayscale} background is the height-map.\label{fig:sc2-img}}
    {\includegraphics[width=0.41\linewidth]{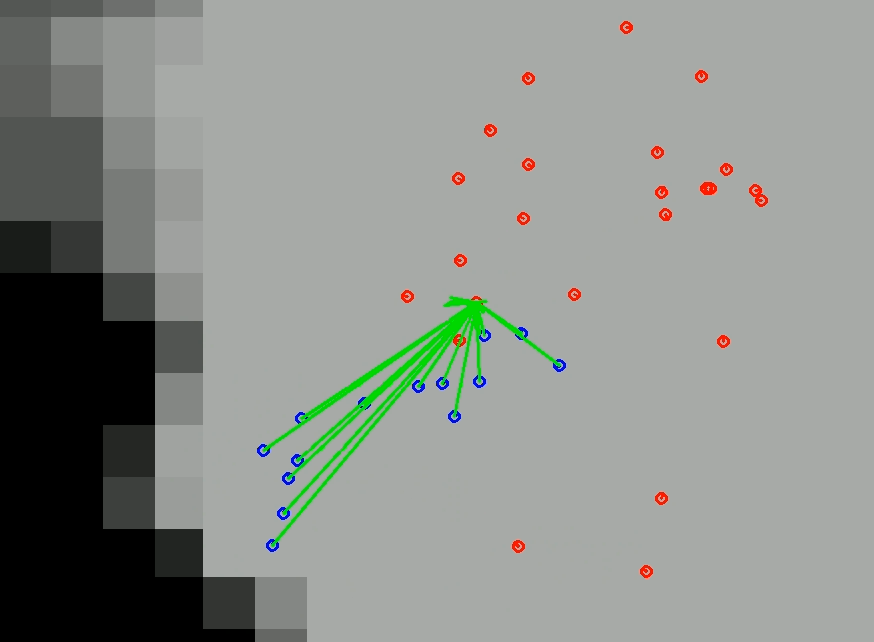}}
    \vspace{-3mm}
    \caption{Visualisation of multi-agent environments used for benchmarking.}
    \label{fig:env-screenshots}
\end{figure}

\subsection{Chasing Targets} \label{sec:env-chasing}

The chasing targets environment involves an arbitrary number of two-wheeled robots that chase an arbitrary number of particles bouncing around the environment, while trying to avoid collisions with one another. Each robot is randomly assigned a target particle at the beginning of the simulation and initialized in a random stationary pose. When the robot reaches the particle, it is randomly assigned to a new particle. Robots are controlled with a simple cost function to select the best control inputs that minimizes the robot's distance to the target's projected position and a penalty term if a collision is forecast to occur with another robot. Invalid control inputs, such as exceeding a speed limit are disallowed. Particles are initialised at random positions in the arena with a random initial velocity sampled from a normal distribution and clamped to a maximum range. A visualisation of the environment is depicted in Figure \ref{fig:gym-img}.

The main modelling objective in this task is to determine the intended target (particle) for each of the robots in the scene. We consider two variations of this task. In the first, the observation space consists of the current pose of the robots and particles. A random number of robots and particles are sampled from a uniform distribution for each training and validation sample. For the second task variation, the observation space only includes the robots, and the model must predict where particles are located in the scene. This is challenging as not only the position of the particles are unknown, but the cardinality must also be estimated. This task is formulated as an occupancy problem, the model needs to estimate the likelihood that a position in the environment is occupied by a particle.

\subsection{StarCraft II} \label{sec:env-sc2}

The StarCraft II data used is sampled from tournaments hosted between 2019 and 2023. To focus on battle sequences where players are micro-managing their units in combat, we target parts of the game when damage dealt or received by a player exceeds a threshold. Non-overlapping sequences of a fixed size are created at these instances. We note that observation data is irregularly sampled, hence the time duration of each sequence will vary. This domain has some subtleties compared to \textit{Chasing-Targets} (Section \ref{sec:env-chasing}) as units are not always assigned to an enemy unit. They can either be idle, or assigned to a position to move to. Hence, we introduce null option for the assignment problem. Furthermore, we introduce a secondary task, where we are required to estimate if a unit has been given a target position command and the location of that target position. The assignment and the position estimation problem are jointly learned and performed by each model.

The observation space for StarCraft II includes a terrain height-map and the position and properties (health, damage, etc.) of player and enemy units. The observation space is restricted to a region-of-interest (ROI) of a fixed size, see Appendix \ref{sec:sc2-roi-calculation} for detail on how the ROI is calculated. An example of an extracted ROI with unit data and their assignments is depicted in Figure \ref{fig:sc2-img}. Units outside of the ROI are truncated, resulting in a time-varying set of units in the environment as they enter, exit or die in combat. The observability rules of the game apply from a player's point-of-view. Enemy units will be hidden in the fog of war or when an obfuscation ability is used, for example Zerg players can ``burrow'' units into the ground.

\begin{wrapfigure}{r}{0.4\textwidth}
    \centering
    \subcaptionbox{\textbf{BERT} Encoder\label{fig:bert}}{\begin{tikzpicture}[thick, ->]

% Vertical Flow
% \node[rectangle, draw, very thick] (cls) {$\sL$};
% \node[rectangle, draw, very thick, right=of cls] (obs){$\sO$};

% \node[fit=(cls)(obs), draw=none, inner sep=0] (fit) {};
% \node[rectangle, draw, very thick, above=of fit, yshift=-0.5cm, minimum width=0.2\linewidth] (encoder) {Transformer Encoder};
% \draw[->] (cls.north) -- (encoder.south -| cls.north);
% \draw[->] (obs.north) -- (encoder.south -| obs.north);

% \node[rectangle, draw, very thick, above=of cls, yshift=0.5cm] (cls_enc) {$\sL'$};
% \draw[->] (encoder.north -| cls.north) -- (cls_enc.south);

% Horizontal Flow
\node[rectangle, draw, very thick] (cls) {$\sL$};
\node[rectangle, draw, very thick, below=of cls] (obs){$\sO$};

\node[fit=(cls)(obs), draw=none, inner sep=0] (fit) {};
\node[rectangle, draw, very thick, right=of fit, rotate=-90, anchor=center] (encoder) {\parbox{2cm}{\centering Transformer\\Encoder}};
\draw[->] (cls.east) -- (encoder.south |- cls.east);
\draw[->] (obs.east) -- (encoder.south |- obs.east);

\node[rectangle, draw, very thick, right=of cls, xshift=1cm] (cls_enc) {$\sL'$};
\draw[->] (encoder.north |- cls_enc.west) -- (cls_enc.west);

% This is the Combined Encoding, but BERT can do piece-wise as well. Combined is used for RNNs
% \node[rectangle, draw, very thick] (cls) {$\sL$};
% \node[rectangle, draw, very thick, right=of cls] (player){$\sO_{player}$};
% \node[rectangle, draw, very thick, right=of player] (enemy){$\sO_{enemy}$};
% \node[rectangle, draw, very thick, below=of player, yshift=0.5cm] (player_e){$\mathbf{E}_{player}$};
% \node[rectangle, draw, very thick, below=of enemy, yshift=0.5cm] (enemy_e){$\mathbf{E}_{enemy}$};

% \node[rectangle, draw, very thick, above=of player, yshift=-0.5cm, minimum width=0.4\linewidth] (encoder) {Transformer Encoder};
% \draw[->] (cls.north) -- (encoder.south -| cls.north);
% \draw[->] (player.north) -- (encoder.south -| player.north);
% \draw[->] (enemy.north) -- (encoder.south -| enemy.north);
% \draw (player_e) -- node[midway, right] {+} (player);
% \draw (enemy_e) -- node[midway, right] {+} (enemy);

% \node[rectangle, draw, very thick, above=of cls, yshift=0.5cm] (cls_enc) {$\sL'$};
% \draw[->] (encoder.north -| cls.north) -- (cls_enc.south);

\end{tikzpicture}}
    \subcaptionbox{\textbf{X-Attn} Encoder\label{fig:xattn}}{\begin{tikzpicture}[thick, ->]

\node[rectangle, draw, very thick] (cls) {$\sL$};

\node[rectangle, draw, very thick, right=of cls, xshift=-0.3cm] (attn) {MHA};
\draw[->] (cls) -- node[midway, above] {$\mathbf{Q}$} (attn);

\node[rectangle, draw, very thick, above=of attn, yshift=-0.3cm] (obs){$\sO$};
\draw[->] (obs) -- node[midway, right] {$\mathbf{K,V}$} (attn);

\node[rectangle, draw, very thick, right=of attn, xshift=-0.3cm] (cls-enc) {$\sL'$};
\draw[->] (attn) -- (cls-enc);

% This is piecewise, cross-attention can do combined as well
% \node[rectangle, draw, very thick, right=of cls] (player-attn) {MHA};
% \draw[->] (cls) -- node[midway, above] {$\mathbf{Q}$} (player-attn);
% \node[rectangle, draw, very thick, above=of player-attn] (player){$\sO_{player}$};
% \draw[->] (player) -- node[midway, right] {$\mathbf{K,V}$} (player-attn);

% \node[rectangle, draw, very thick, right=of player-attn] (enemy-attn) {MHA};
% \draw[->] (player-attn) -- node[midway, above] {$\mathbf{Q}$} (enemy-attn);
% \node[rectangle, draw, very thick, above=of enemy-attn] (enemy){$\sO_{enemy}$};
% \draw[->] (enemy) -- node[midway,right] {$\mathbf{K,V}$} (enemy-attn);

% \node[rectangle, draw, very thick, right=of enemy-attn] (cls-enc) {$\sL'$};
% \draw[->] (enemy-attn) -- (cls-enc);

\end{tikzpicture}}
    \caption{Encoders summarize an irregular set of tokens from the observation $\sO$, to a fixed size $\sL$, for the spatio-temporal encoder.}
    \label{fig:encoder-types}
    \vspace{-20mm}
\end{wrapfigure}
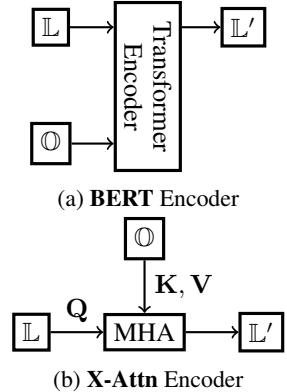

Both environments above are good examples of sequential modelling and prediction problems involving irregular, multidimensional time varying observation spaces. Below, we describe a sequential encoding and dynamics modelling strategy suitable for these tasks.

\section{Method} \label{sec:method}

We evaluate several methods to encode spatio-temporal features into a series of latent states. The decoder for each of the tasks is fixed in order to isolate the contribution of the encoding methodology.

\subsection{Spatio-Temporal Encoding} \label{sec:method:encode}

This section outlines how the latent representation of the scene is constructed, using the StarCraft task above as an example. The scene observation is tokenized identically for each of these encoders. First, the $(x,y,\theta)$ pose of the agents are sinusoidally encoded. Units from StarCraft II have additional information included such as health and a learned embedding associated with the type of unit. From this variably sized set of tokens, $\sO'=\{\vo\}_{n_o}, \vo\in\R^{d_o}$, sampled from the observation, we must summarize a fixed set $\sL = \{\vl\}_{n_l}, \vl\in\R^{d_l}$ suitable for latent dynamics modelling.

We use two methods for summarizing $\sO'$ into $\sL'$, as illustrated in Figure \ref{fig:encoder-types}, each of which involves Multi-head Attention between $\sO'$ and $\sL$. The first is a \textbf{BERT}-style \citep{devlin2018bert} transformer encoder. Latents $\sL$ are concatenated to $\sO'$ , and act as the ``[CLS]'' tokens of the BERT encoder. The encoded ``[CLS]'' tokens, $\sL'$, are used as a fixed-size representation of the variably sized observation data. The second method uses a block of cross-attention layers (\textbf{X-Attn}), $\sL$ to query key-value pairs generated from $\sO'$ to transfer relevant information. We denote these encoders as $\mathbf{Enc}$.

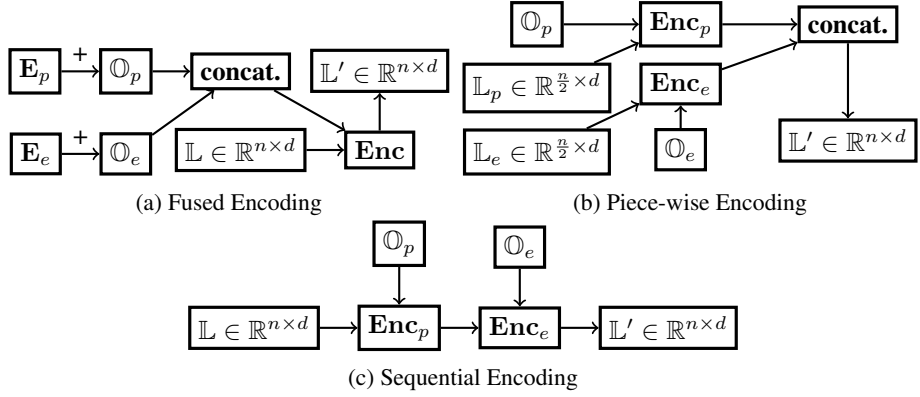
\begin{figure}
    \centering
    \subcaptionbox{Fused Encoding\label{fig:enc-fused}}{\begin{tikzpicture}[thick, ->]

\node[rectangle, draw, very thick] (player){$\sO_{p}$};
\node[rectangle, draw, very thick, left=of player, xshift=0.5cm] (player_e){$\mathbf{E}_{p}$};
\draw (player_e) -- node[midway, above] {+} (player);

\node[rectangle, draw, very thick, below=of player, yshift=0.5cm] (enemy){$\sO_{e}$};
\node[rectangle, draw, very thick, left=of enemy, xshift=0.5cm] (enemy_e){$\mathbf{E}_{e}$};
\draw (enemy_e) -- node[midway, above] {+} (enemy);

\node[rectangle, draw, very thick, right=of player, xshift=-0.5cm] (cat) {\textbf{concat.}};
\draw[->] (player) -- (cat);
\draw[->] (enemy) -- (cat);

\node[rectangle, draw, very thick, below=of cat, yshift=0.5cm] (cls) {$\sL\in\R^{n\times d}$};

\node[rectangle, draw, very thick, right=of cls, xshift=-0.5cm] (encoder) {$\mathbf{Enc}$};
\draw[->] (cls) -- (encoder);
\draw[->] (cat) -- (encoder);

\node[rectangle, draw, very thick, above=of encoder, yshift=-0.5cm] (cls_enc) {$\sL'\in\R^{n\times d}$};
\draw[->] (encoder) -- (cls_enc);

\end{tikzpicture}}
    \subcaptionbox{Piece-wise Encoding\label{fig:enc-piecewise}}{\begin{tikzpicture}[thick, ->]

\node[rectangle, draw, very thick] (player){$\sO_{p}$};
\node[rectangle, draw, very thick, below=of player, yshift=0.8cm] (player-latent) {$\sL_{p}\in\R^{\frac{n}{2}\times d}$};
\node[rectangle, draw, very thick, right=of player] (player-enc) {$\mathbf{Enc}_{p}$};
\draw[->] (player-latent) -- (player-enc);
\draw[->] (player) -- (player-enc);

\node[rectangle, draw, very thick, below=of player-enc, yshift=0.8cm] (enemy-enc) {$\mathbf{Enc}_{e}$};
\node[rectangle, draw, very thick, below=of player-latent, yshift=0.8cm] (enemy-latent) {$\sL_{e}\in\R^{\frac{n}{2}\times d}$};
\node[rectangle, draw, very thick, below=of enemy-enc, yshift=0.7cm] (enemy){$\sO_{e}$};
\draw[->] (enemy-latent) -- (enemy-enc);
\draw[->] (enemy) -- (enemy-enc);

\node[rectangle, draw, very thick, right=of player-enc] (cat) {\textbf{concat.}};
\node[rectangle, draw, very thick, below=of cat] (out) {$\sL'\in\R^{n\times d}$};
\draw[->] (player-enc) -- (cat);
\draw[->] (enemy-enc) -- (cat);
\draw[->] (cat) -- (out);

% \node[rectangle, draw, very thick] (player-latent) {$\sL_{p}\in\R^{\frac{n}{2}\times d}$};
% \node[rectangle, draw, very thick, below=of player-latent, yshift=0.8cm] (player){$\sO_{p}$};

% \node[fit=(player)(player-latent), draw=none, inner sep=0] (player-fit) {};
% \node[rectangle, draw, very thick, right=of player-fit] (player-enc) {$\mathbf{Enc}_{p}$};
% \draw[->] (player-latent) -- (player-enc);
% \draw[->] (player) -- (player-enc);

% \node[rectangle, draw, very thick, below=of player, yshift=0.8cm] (enemy-latent) {$\sL_{e}\in\R^{\frac{n}{2}\times d}$};
% \node[rectangle, draw, very thick, below=of enemy-latent, yshift=0.8cm] (enemy){$\sO_{e}$};

% \node[fit=(enemy)(enemy-latent), draw=none, inner sep=0] (enemy-fit) {};
% \node[rectangle, draw, very thick, right=of enemy-fit] (enemy-enc) {$\mathbf{Enc}_{e}$};
% \draw[->] (enemy-latent) -- (enemy-enc);
% \draw[->] (enemy) -- (enemy-enc);

% \node[rectangle, draw, very thick, right=of player-enc] (cat) {\textbf{concat.}};
% \node[rectangle, draw, very thick, below=of cat] (out) {$\sL'\in\R^{n\times d}$};
% \draw[->] (player-enc) -- (cat);
% \draw[->] (enemy-enc) -- (cat);
% \draw[->] (cat) -- (out);

\end{tikzpicture}}
    \subcaptionbox{Sequential Encoding\label{fig:enc-seq}}{\begin{tikzpicture}[thick, ->]
\node[rectangle, draw, very thick] (cls) {$\sL\in\R^{n\times d}$};
\node[rectangle, draw, very thick, right=of cls, xshift=-0.5cm] (player-attn) {$\mathbf{Enc}_p$};
\draw[->] (cls) -- (player-attn);
\node[rectangle, draw, very thick, above=of player-attn, yshift=-0.5cm] (player){$\sO_{p}$};
\draw[->] (player) -- (player-attn);

\node[rectangle, draw, very thick, right=of player-attn, xshift=-0.5cm] (enemy-attn) {$\mathbf{Enc}_e$};
\draw[->] (player-attn) -- (enemy-attn);
\node[rectangle, draw, very thick, above=of enemy-attn, yshift=-0.5cm] (enemy){$\sO_{e}$};
\draw[->] (enemy) -- (enemy-attn);

\node[rectangle, draw, very thick, right=of enemy-attn, xshift=-0.5cm] (cls-enc) {$\sL'\in\R^{n\times d}$};
\draw[->] (enemy-attn) -- (cls-enc);
\end{tikzpicture}}
    \caption{Several methods of encoding player ($\sO_p$) and enemy ($\sO_e$) observations to a fixed dimension $\sL\in\R^{n\times d}$. Process together with an embedding to distinguish $\sO_p$ from $\sO_e$ (Fig. \ref{fig:enc-fused}), process separately (Fig. \ref{fig:enc-piecewise}) or process sequentially (Fig. \ref{fig:enc-seq}).}
    \label{fig:combine-types}
\end{figure}

Without loss of generality, consider the case where we have two observation sources. In the tasks above, these correspond to the two teams of agents. Since there are two distinct observation sources, $\sO_p$ and $\sO_e$, we test a variety of methods to determine an effective method for combining both into $\sL'$. The ``Fused'' method, Figure \ref{fig:enc-fused}, adds a learned embedding per source and then concatenates the sources together for encoding. This method has fewer parameters than the alternative methods as there is only one $\mathbf{Enc}$. This also enables flexibility in gathering the optimal amount of information from each type of source. However, this comes at the cost of a larger attention matrix inside $\mathbf{Enc}$ with $\mathcal{O}((N_{p} + N_{e})^2)$. The ``Piece-wise'' method, Figure \ref{fig:enc-piecewise}, encodes $\sO_p$ and $\sO_e$ separately with half of the latent state $\sL$ used for each observation source. This method has the benefit of a smaller attention matrix $\mathcal{O}(N_{p}^2 + N_{e}^2)$, but enforces an equal weighting of information between the two sources, which is potentially sub-optimal. ``Sequental'' processes $\sO_p$ then $\sO_e$, as depicted in Figure \ref{fig:enc-seq}. This renders a smaller attention matrix, but reduces parallelism and increases model depth.

\subsubsection{Scan Encoder} \label{sec:method:enc:iscan}

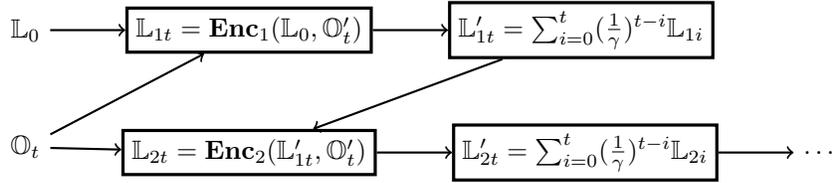
\begin{figure}[hbt]
    \centering
    \begin{tikzpicture}[thick, ->]
\node (init) {$\sL_0$};
\node[below=of init] (obs) {$\sO_t$};
\node[rectangle, draw, very thick, right=of init] (enc1) {$\sL_{1t}=\mathbf{Enc}_1(\sL_0, \sO_t')$};
\node[rectangle, draw, very thick, right=of enc1] (scan1) {$\sL_{1t}' = \sum_{i=0}^{t} (\frac{1}{\gamma})^{t-i}\sL_{1i}$};

\node[rectangle, draw, very thick, below=of enc1] (enc2) {$\sL_{2t} =\mathbf{Enc}_2(\sL_{1t}', \sO_t')$};
\node[rectangle, draw, very thick, right=of enc2] (scan2) {$\sL_{2t}' = \sum_{i=0}^{t} (\frac{1}{\gamma})^{t-i}\sL_{2i}$};

\node[rectangle, right=of scan2] (dots) {$\dots$};

\draw (init) edge (enc1) (obs) edge (enc1);
\draw (enc1) edge (scan1);
\draw (scan1) edge (enc2) (obs) edge (enc2);
\draw (enc2) edge (scan2);
\draw (scan2) edge (dots);
\end{tikzpicture}
    \caption{The inclusive-scan encoder alternates between sampling the observation based on some latent variable $\sL_x$ and accumulating a weighted sum $\sL'$. A learned set of parameters $\sL_0$ is used as the query in the first block. $\gamma\geq1$}
    \label{fig:enc-iscan}
\end{figure}

The key design objective of this spatio-temporal encoder is to efficiently aggregate historical information with a parallel algorithm, and to use that accumulated knowledge to sample the current observation. Furthermore, we use a set of tokens to represent our hidden state, rather than a single vector that is common to most algorithms. The motivation behind this is that attention mechanisms can be then utilized effectively with this hidden state, to either inject or extract information from this state. To achieve this, we use a weighted sum of an initially encoded input data from previous steps in the sequence. The accumulation can be performed efficiently in parallel over the sequence dimension with an inclusive scan operation \cite{merrill2016scan}. The proposed inclusive-scan algorithm uses the cross-attention encoder (Fig. \ref{fig:xattn}) with the sequential algorithm (Fig. \ref{fig:enc-seq}) to create a recursive query driven sequential modelling approach. Inclusive-scan is performed on the updated variables to accumulate temporal information. This allows variables to accumulate context, so that when an observation is re-queried, it is conditioned on an accumulated history. This process is depicted in Figure \ref{fig:enc-iscan}. At inference time, we can calculate the next accumulated encoding with a scaled copy of the previous time-step $\sL_{xt}'=\frac{1}{\gamma}\sL_{x(t-1)}' + \sL_{xt}$, resulting in a computational complexity of $\mathcal{O}(1)$ with respect to the length of the sequence. We show in Section \ref{sec:results} that cycling between cross-attention and inclusive-scan is superior to a block of cross-attention layers with an inclusive-scan at the end or sampling the input once, and cycling between inclusive-scan and self-attention. We include the number of cycles as nomenclature for this scan encoder in Section \ref{sec:results}, for example Scan $4\times$ is a scan encoder with four cycles.

An important subtlety is that the inclusive-scan is weighted such that the historical contribution decays over time with $\gamma\geq1$. This ensures that the accumulation does not diverge in magnitude. We show in Section \ref{sec:results} that the scan with $\gamma=2$ outperforms $\gamma=1$ (a simple cumulative sum). A PyTorch extension was written to efficiently perform the forward and backward method of discounted inclusive-scan both on CPU and on GPU with CUDA\footnote{\url{https://github.com/5had3z/torch-discounted-cumsum-nd}}.

\subsubsection{Baseline Encoders} \label{sec:method:enc:baseline}
% \subsubsection{Recursive Transformer} \label{sec:method:encode:rec-tf}
% The \textbf{recursive transformer} (Recursive) uses a latent-variable query approach to transfer observation information to the latent state representation, akin to the inclusive-scan algorithm. At each new observation, cross-attention is incrementally applied between the latent variables and the units and then the targets. This incremental processing is efficient at evaluation time, however results in long gradient chains during training, significantly slowing the training process in comparison to other algorithms.

% \subsubsection{Spatio-Temporal Transformer} \label{sec:method:encode:st-tf}
The \textbf{spatio-temporal transformer} (S.T.) groups tokens from the same time-step to create observation chunks at each time-step. A learned position embedding of the absolute time-step is added to the tokens. We use a transformer encoder with causal masking to process the entire sequence. We evaluate both the BERT and cross-attention encoders on the agent assignment challenge.

% \subsubsection{Residual Neural Networks} \label{sec:method:encode:rnn}
We evaluate each of the \textbf{recurrent neural networks} included in PyTorch: RNN, GRU and LSTM. We use BERT to encode the observation data and a set of recurrent modules to process each latent encoding. We find that learned initialization performs better than zero initialization with an LSTM and therefore also use learned initialization for other module structures (Appendix \ref{sec:rnn-compare}).

We use \textbf{Mamba2} as a point of comparison against modern SSMs. The cross-attention encoder is used to encode observation data, and like the recurrent neural networks above, we use a set of Mamba2 modules to process each latent encoding. We use 64 for the state dimension, 4 for the convolution dimension and an expansion factor of 2 as the parameters for Mamba2. 

% \subsubsection{Height-Map Encoder} \label{sec:method:encode:mini-enc}

The height-map is additional context provided in the SC2 task. ResNet-18 is used as the feature extractor, however we replace the final $1\times1$ adapative-average-pool and fully-connected layers with a $4\times4$ adaptive-average-pool layer to create a grid of features. Sinusoidal position embeddings are added to the feature grid and then flattened, creating the set of contextual tokens. These tokens are appended to the extracted spatio-temporal features before being passed to the task decoder.

\subsection{Task Decoding}

In this section, we detail how the decoding strategy used for each benchmark task. The foundation of each task decoder is a cross-attention operation to gather information from the latent state.

\textbf{Goal Assignment} aims to find the correlation between agents and targets. To encode the agent, we first project the agent feature vector with a linear layer, then use cross-attention to transfer information from the latent state that is useful to associate agents and targets. The targets are also projected with a linear layer. For the StarCraft II challenge, a learned null token is appended to the set of projected targets. Finally, the correlation between agents and targets is determined using a dot-product. Soft-max is applied along the target dimension to calculate a final weighting between the agents and targets. We note that this is a one-to-many problem, many agents can have the same target, but each agent only has one target (including the null assignment).

We utilize the negative log likelihood (NLL) as a categorical loss between the predicted assignment and the ground truth assignment. However, there are some complications due to padding introduced to make regular tensor shapes from the variable agents and targets in the scene. Invalid agents can simply be omitted from the loss function with a mask. To handle invalid targets, we group agents with the same number of targets in the scene and truncate the tensor to this number. This allows us to apply NLL loss while avoiding gradients from invalid targets and agents. In the StarCraft II task, there is a significant imbalance between agents with a target and those without. A weighting factor of $0.05$ is applied to emphasise non-null assignment losses.

A \textbf{Movement Target Assignment} task completes the motion action space that units in StarCraft II can take. Here, the objective is to estimate the coordinates on the map where the unit is moving to. This plays an important part of the game as manoeuvring units into advantageous positions such as the high ground or choke-points is an important strategy in game-play. We evaluate several methods of estimating the target position with both regression and categorical techniques including: global cartesian, relative cartesian and relative polar. Furthermore, a logit is emitted to estimate the likelihood that the unit is following a position command and position estimate is valid.

An L2 loss is used for regression-based position estimation and HL-Gauss \citep{pmlr-v80-imani18a} is used for categorical-based estimation. We include additional logic in the HL-Gauss loss to correct angle wrapping for polar coordinate estimation, further detail is provided in Appendix \ref{sec:sc2-position}. Binary cross-entropy loss is used for the valid position command logit. Position commands are under-represented in the dataset compared to target commands. To increase the amount of position data, we create faux position commands from the coordinates of target units. This does not effect the valid position command logit training.

\textbf{Hidden Target Estimation} in \textit{Chasing-Targets} is formulated as an occupancy prediction problem. An array of coordinates is sinusoidally encoded to query the spatio-temporal encoding the likelihood of hidden target occupation. This is performed with cross-attention between the positional query and the spatio-temporal encoding. A linear layer transforms this result to a single logit. Focal loss \cite{focal} is used for occupancy prediction to address class imbalance between unoccupied and occupied pixels. This rendering method can create an image of any resolution as position queries can be continuous, however we use a fixed grid for simplicity.

\section{Results} \label{sec:results}

We used PyTorch to train our models. Unless otherwise specified, experiments used a batch size of 64, the AdamW optimizer with a learning rate of $1e^{-4}$ and a polynomial schedule with power 0.9 and gradient clipping of 0.1\footnote{Link to source and model configurations will be provided later}.

\subsection{Chasing Targets}

Each instance of the environment is randomly generated each time, sampling the number of robots and targets from a uniform distribution, and randomly placing them on the field. The field is a $4\times4$m grid and the maximum velocity of the agents is 0.5m/s. The first 10 iterations of the simulation are skipped to remove the domain gap between the behaviour of the robots after random initialization, and the steady state chasing and switching targets. The duration of the simulation is 41 iterations.

\subsubsection{Target Assignment}

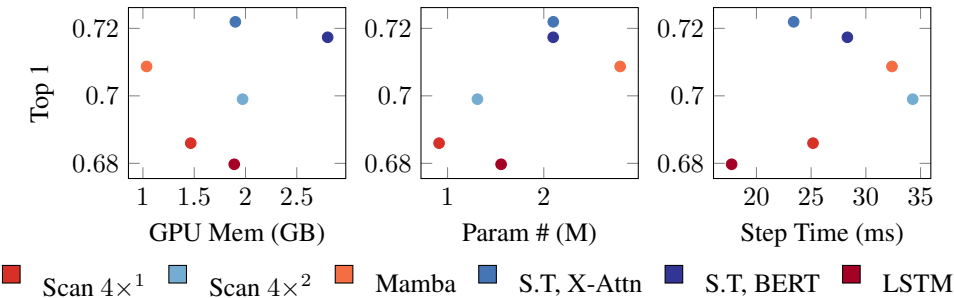
\begin{figure}
    \centering
    \begin{tikzpicture}
    \begin{groupplot}[
        width=0.32\textwidth,
        group style={
            group name=gym-results,
            group size=3 by 1,
            xlabels at=edge bottom,
            ylabels at=edge left,
            horizontal sep=1cm,
            vertical sep=1cm,
        },
        ylabel={Top 1},
        scatter,
        point meta=explicit symbolic,
        visualization depends on=\thisrow{label} \as \label,
        scatter/classes={
            {Scan 4-1}={mark=*,lightred},
            {Scan 4-2}={mark=*,lightblue},
            {Mamba}={mark=*,darkorange},
            {S.T w/ X-Attn}={mark=*,mediumblue},
            {S.T w/ BERT}={mark=*,darkblue},
            {LSTM}={mark=*,darkred}
        },
        % scaled ticks=false,
        % legend to name=sharedlegend,
        legend style={at={(1.05,1)}, anchor=north west}
    ]

    \nextgroupplot[xlabel={GPU Mem (GB)}]
    \addplot[
        only marks,
        scatter src=explicit symbolic,
    ] table[meta=label] {
        x y label
        1.465931264 0.6859842160852944 {Scan 4-1}
        1.971753472 0.6990119189750857 {Scan 4-2}
        1.035480576 0.708710228524557 {Mamba}
        1.901103616 0.7219344813649248 {S.T w/ X-Attn}
        2.801005568 0.7173531695110041 {S.T w/ BERT}
        1.890305536 0.679722018358184 {LSTM}
    };
    
    \nextgroupplot[xlabel={Param \# (M)}]
    \addplot[
        only marks,
        scatter src=explicit symbolic
    ] table[meta=label] {
        x y label
        0.911808 0.6859842160852944 {Scan 4-1}
        1.310144 0.6990119189750857 {Scan 4-2}
        2.793792 0.708710228524557 {Mamba}
        2.098496 0.7219344813649248 {S.T w/ X-Attn}
        2.097472 0.7173531695110041 {S.T w/ BERT}
        1.556224 0.679722018358184 {LSTM}
    };

    \nextgroupplot[xlabel={Step Time (ms)}]
    \addplot[
        only marks,
        scatter src=explicit symbolic,
    ] table[meta=label] {
        x y label
        25.171627319650725 0.6859842160852944 {Scan 4-1}
        34.28227559197694 0.6990119189750857 {Scan 4-2}
        32.38541091923253 0.708710228524557 {Mamba}
        23.402895207254908 0.7219344813649248 {S.T w/ X-Attn}
        28.312769798503723 0.7173531695110041 {S.T w/ BERT}
        17.733362732542453 0.679722018358184 {LSTM}
    };

    \end{groupplot}

    % Manually create the legend with correct colors and move it below the plot
    \matrix[anchor=north, column sep=5pt, row sep=3pt] at (current bounding box.south) {
        \node[draw, fill=lightred, mark options={lightred}, mark=*] at (0,0) {}; & \node {Scan $4\times^1$}; &
        \node[draw, fill=lightblue, mark options={lightblue}, mark=*] at (0,0) {}; & \node {Scan $4\times^2$}; &
        \node[draw, fill=darkorange, mark options={darkorange}, mark=*] at (0,0) {}; & \node {Mamba}; &
        \node[draw, fill=mediumblue, mark options={mediumblue}, mark=*] at (0,0) {}; & \node {S.T, X-Attn}; &
        \node[draw, fill=darkblue, mark options={darkblue}, mark=*] at (0,0) {}; & \node {S.T, BERT}; &
        \node[draw, fill=darkred, mark options={darkred}, mark=*] at (0,0) {}; & \node {LSTM}; \\
    };
\end{tikzpicture}

% \definecolor{darkred}{RGB}{165,0,38}
% \definecolor{lightred}{RGB}{215,48,39}
% \definecolor{darkorange}{RGB}{244,109,67}
% \definecolor{lightorange}{RGB}{253,174,97}
% \definecolor{lightblue}{RGB}{116,173,209}
% \definecolor{mediumblue}{RGB}{69,117,180}
% \definecolor{darkblue}{RGB}{49,54,149}
    \caption{Average Top 1 assignment accuracy over the simulation sequence with training cost. The top left corner is ideal in each scenario. The Scan encoder linearly scales in performance and cost from $^1$ to $^2$ by introducing an additional self-attention layer in each recursion block.}
    \label{fig:gym-acc-cost}
\end{figure}

We train the target assignment challenge for $\approx47$k iterations and use a latent state $\mathcal{L}\in\R^{8\times128}$. The number of robots in each simulation is sampled from $\mathcal{U}(8,15)$ and the number of targets $\mathcal{U}(3,6)$. Figure \ref{fig:gym-acc-cost} shows the complex trade-space between training throughput, memory usage, and the parameter count of the model. The LSTM is the fastest model to train with a step time of 17.7ms, has the second lowest parameter count and third lowest memory consumption. The most accurate model, the spatio-temporal (S.T) transformer with X-Attn encoder has the second shortest step time, moderate memory consumption and parameter count. The Scan $\times 4$ encoder has the least number of parameters by a large margin, however suffers from lower accuracy in this task compared to S.T and Mamba2. Mamba2 performs similarly to S.T with the lowest memory consumption during training.

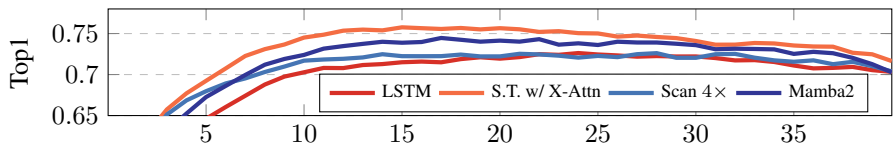
\begin{figure}
    \centering
    \begin{tikzpicture}
\begin{axis}[
    height=3cm,
    width=12cm,
    ylabel={Top1},
    xmin=0, xmax=40,
    ymin=0.65, ymax=0.78,
    xtick={5,10,15,20,25,30,35},
    ytick={0.65,0.7,0.75},
    legend columns=4,
    legend pos=south east,
    legend style={
        font=\scriptsize,
        legend cell align=left,
    },
    ymajorgrids=true,
    grid style=dashed,
    every axis plot/.append style={ultra thick},
    ]
\addplot[color=lightred] table [x index=0, y index=1, col sep=comma]{data/timeseries-perf.csv};
\addlegendentry{LSTM}
\addplot[color=darkorange] table [x index=0, y index=2, col sep=comma]{data/timeseries-perf.csv};
\addlegendentry{S.T. w/ X-Attn}
\addplot[color=mediumblue] table [x index=0, y index=3, col sep=comma]{data/timeseries-perf.csv};
\addlegendentry{Scan $4\times$}
\addplot[color=darkblue] table [x index=0, y index=4, col sep=comma]{data/timeseries-perf.csv};
\addlegendentry{Mamba2}
\end{axis}
\end{tikzpicture}
    \caption{Assignment accuracy comparison over a sequence.}
    \label{fig:accuarcy-time}
\end{figure}

We notice a trend of assignment accuracy decay after a peak earlier in the sequence in Figure \ref{fig:accuarcy-time}. This is more pronounced in the better performing S.T and Mamba2 models, converging to approximately the same performance as the Scan $4\times$ and LSTM by the end of the sequence.

To understand the contributions of each of the design decisions in the Scan encoder, we remove individual components from a baseline model. If a cumulative sum is performed without weighting, accuracy decays towards the end of the sequence, and in general underperforms the baseline $\gamma=2$. Reducing the size of the latent state to $\R^{1\times128}$ has significant performance consequences as not enough historical information is retained over the sequence. This performs almost as poorly as not performing a scan at all, the worst performing experiment conducted.
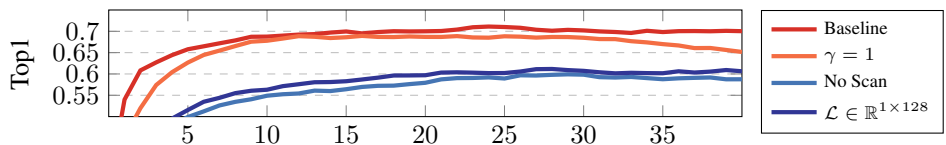
\begin{figure}
    \centering
    \begin{tikzpicture}
\begin{axis}[
    height=3cm,
    width=10cm,
    ylabel={Top1},
    xmin=0, xmax=40,
    ymin=0.5, ymax=0.75,
    xtick={5,10,15,20,25,30,35},
    ytick={0.55,0.6,0.65,0.7},
    legend columns=1,
    legend pos=outer north east,
    legend style={
        font=\scriptsize,
        legend cell align=left,
    },
    ymajorgrids=true,
    grid style=dashed,
    every axis plot/.append style={ultra thick},
    ]
\addplot[color=lightred] table [x index=0, y index=3, col sep=comma]{data/scan-ablation.csv};
\addlegendentry{Baseline}
\addplot[color=darkorange] table [x index=0, y index=1, col sep=comma]{data/scan-ablation.csv};
\addlegendentry{$\gamma=1$}
\addplot[color=mediumblue] table [x index=0, y index=2, col sep=comma]{data/scan-ablation.csv};
\addlegendentry{No Scan}
\addplot[color=darkblue] table [x index=0, y index=4, col sep=comma]{data/scan-ablation.csv};
\addlegendentry{$\mathcal{L}\in\R^{1\times128}$}
\end{axis}
\end{tikzpicture}
    \caption{Contribution ablation for a Scan $2\times$ encoder over a sequence.}
    \label{fig:scan-ablate}
\end{figure}

\subsubsection{Hidden Target Estimation}

\begin{figure}
    \centering
    \begin{tikzpicture}
\begin{axis}[
    height=3.5cm,
    width=12cm,
    ylabel={AUC},
    xmin=0, xmax=40,
    ymin=0.0, ymax=0.22,
    xtick={5,10,15,20,25,30,35},
    ytick={0.05,0.1,0.15,0.2},
    legend columns=4,
    legend pos=north west,
    legend style={
        font=\scriptsize,
        legend cell align=left,
    },
    yticklabel style={
        /pgf/number format/fixed,
        /pgf/number format/precision=2
    },
    scaled y ticks=false,
    ymajorgrids=true,
    grid style=dashed,
    every axis plot/.append style={ultra thick},
    ]
\addplot[color=lightred] table [x index=0, y index=1, col sep=comma]{data/val-gym-targets_AUC.csv};
\addlegendentry{Scan $4\times$}
\addplot[color=darkorange] table [x index=0, y index=2, col sep=comma]{data/val-gym-targets_AUC.csv};
\addlegendentry{S.T. w/ X-Attn}
\addplot[color=mediumblue] table [x index=0, y index=3, col sep=comma]{data/val-gym-targets_AUC.csv};
\addlegendentry{Mamba2}
\end{axis}
\end{tikzpicture}
    \caption{Model prediction AUC over the hidden target estimation sequence.}
    \label{fig:uncover-results}
\end{figure}
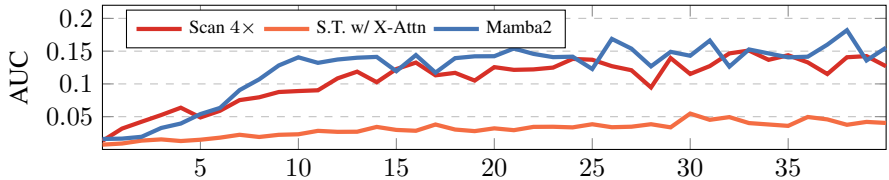

While the S.T with cross attention performs well in target assignment accuracy, it struggles with hidden target estimation. We train the hidden target estimation challenge for $\approx94$k iterations with a batch size of 32 and latent state $\mathcal{L}\in\R^{16\times128}$. The number of robots in each simulation is sampled from $\mathcal{U}(4,12)$ and the number of targets from $\mathcal{U}(2,5)$. Figure \ref{fig:uncover-results} shows that the Scan and Mamba encoders perform equivalently, whereas the transformer encoder fails on this task.

\begin{figure}
    \centering
    \subcaptionbox{0}{\includegraphics[width=0.30\linewidth]{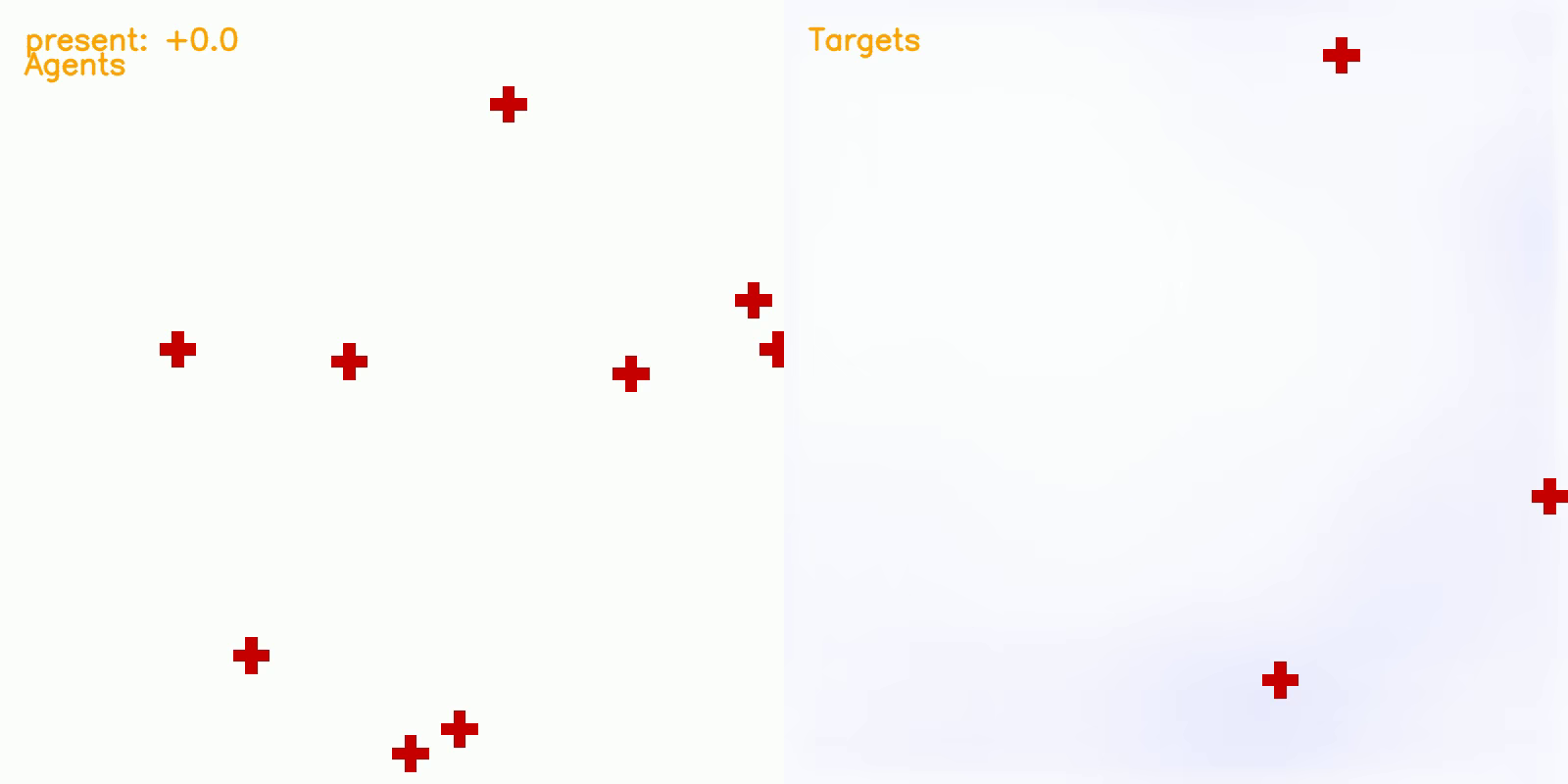}}
    % \hfill
    % \subcaptionbox{10}{\includegraphics[width=0.18\linewidth]{images/uncover-seq/10.png}}
    \hfill
    \subcaptionbox{20}{\includegraphics[width=0.30\linewidth]{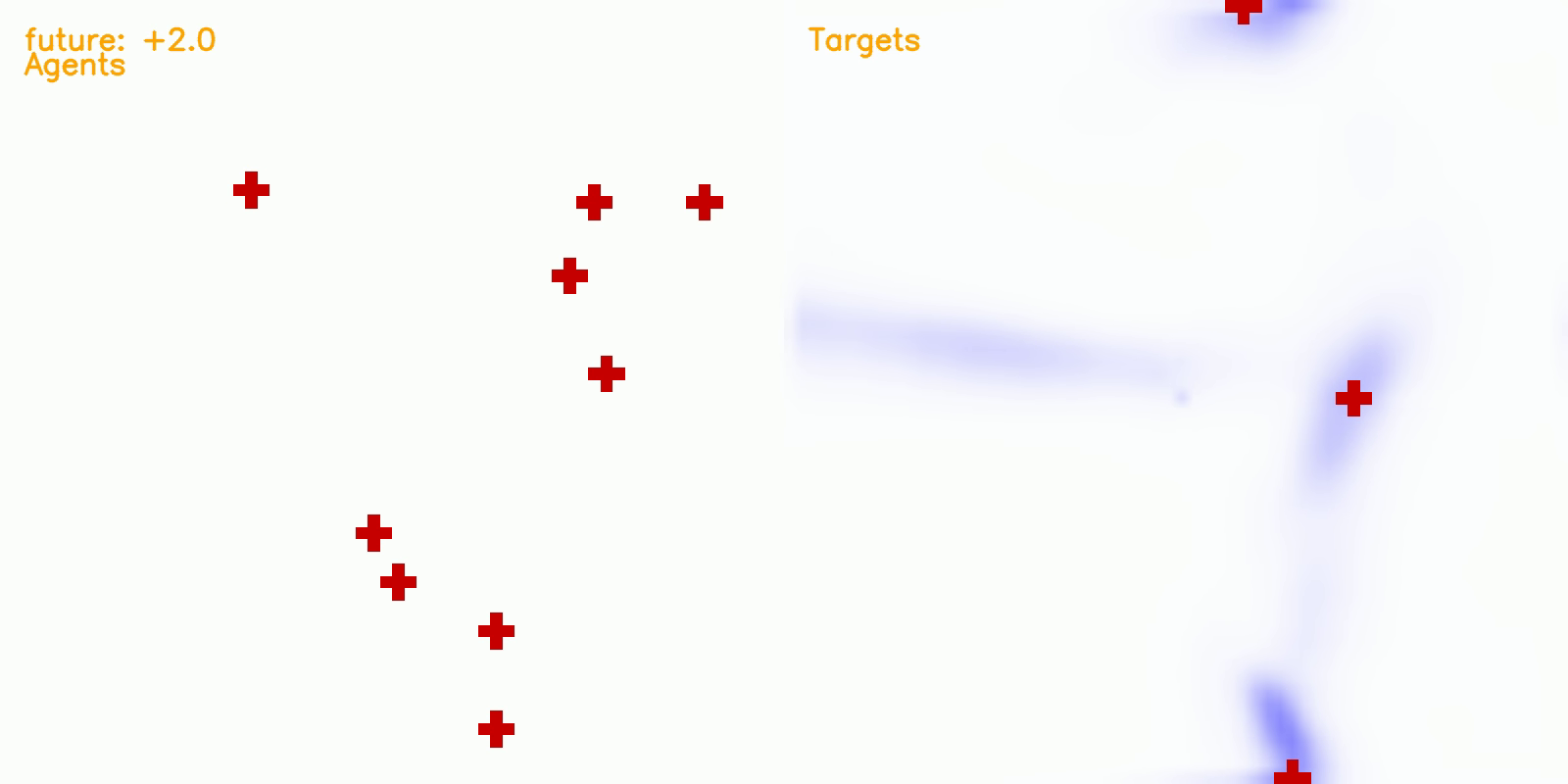}}
    % \hfill
    % \subcaptionbox{30}{\includegraphics[width=0.18\linewidth]{images/uncover-seq/30.png}}
    \hfill
    \subcaptionbox{40}{\includegraphics[width=0.30\linewidth]{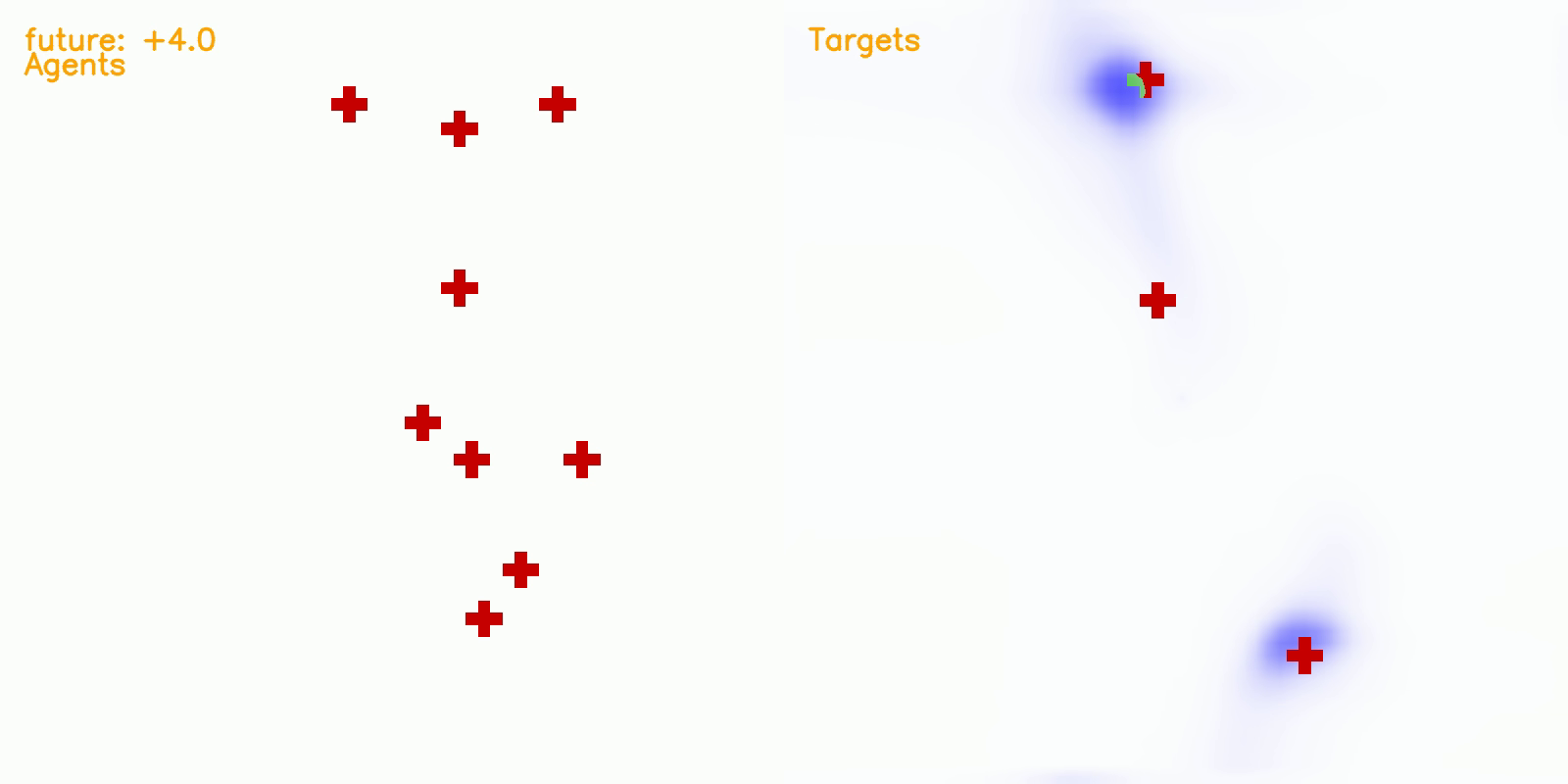}}
    \caption{Rendered hidden target estimation over a 40 frame sequence where the left of each frame is the observable robots and the right is the hidden targets and model prediction. In the right frames, \textcolor{blue}{blue} is the prediction and its intensity represents confidence, \textcolor{red}{red} pixels are the robots and targets.}
    \label{fig:gym-target-est}
\end{figure}

Figure \ref{fig:gym-target-est} illustrates a prediction sequence from this task. Although numerically Scan and Mamba produce low AUC scores in absolute terms, they are often correctly able to infer target locations in the scene.

\subsection{StarCraft II}

We train each model for $\approx205$k iterations with a battle sequence length of 30 and use latent state $\mathcal{L}\in\R^{16\times256}$. StarCraft II is more challenging than \textit{Chasing-Targets}, evident by the significantly lower top-1 assignment score for non-null assignments in Table \ref{tab:sc2-perf}. There are substantially more units on the field, which also vary in cardinality as they enter, exit and die in combat. Furthermore, in dense combat scenarios it is potentially more challenging for the model to precisely assign targets. We find that relative cartesian categorical representation of position performs the best and is used for the following experiments. An ablation of position representation can be found in Appendix \ref{sec:sc2-position}.

There is change in ranking of training cost from \textit{Chasing-Targets} to SC2, as the duration of the sequence is shorter by 25\%, and the task results in a difference in the interplay between efficient sampling of the observation, and temporal accumulation. As the number of units on the screen is significantly greater than \textit{Chasing-Targets}, quickly compressing the observation space with the shallowest model is more important. The baseline encoders use two block repetitions for encoding the observation, compared to Scan encoder which has encoding blocks correlated with the number of repetitions, four in the case of Scan $4\times$. A larger latent state is used as the complexity of the scene and task has increased. All encoders increase in parameter count with the token dimension. Adding more tokens to the latent state increases in number of LSTM and Mamba2 modules. 

% LSTM          8db42aac85585e6b252129eeccda3296 - unit-enc=lstm;latent=16x256;h0=learn
% S.T w/ X-Attn e7c334dcbb6c7b829f05026dd3223751 - unit-enc=tformer;latent=16x256;unit-enc=dec
% Mamba2        57aef86e8d39639db64898da186772f4 - unit-enc=mamba;latent=16x256;unit-enc=dec
% Recursive     d52c9e41f6467ae58e40172c30e85ec5 - unit-enc=v1;latent=16x256;roi=32;+combine-hs
% Scan 6x       9eadfb71e17e09ffcf5fa4539c0a8d57 - unit-enc=v2;latent=16x256;roi=32;+combine-hs
% Scan 4x       90bc8332dda3e32e718bf2598906f083 - unit-enc=v2;latent=16x256;roi=32;block=4
% Scan 4x -sa   88bb169d85fa0ea5b67c293761071e79 - unit-enc=v2;latent=16x256;block=4;sattn=0

\begin{table}[]
    \centering
    \begin{tabular}{cccccccccc}
\multirow{2}{*}{Encoder} &\multirow{2}{*}{Top1}&Top1&\multirow{2}{*}{Top5}&\multirow{2}{*}{MSE}&\multirow{2}{*}{F1}&\# M& Mem&Train\\
                         &                     &+null&                   &                    &                    &Param.&GB&it/sec\\
\hline
        LSTM                &\textbf{0.271}&0.899&\textbf{0.656}&0.200&0.793& 22.1 & 17.5  & 9.00 \\
        S.T w/ X-Attn   &0.264&0.906&0.656&0.191&0.809& \textbf{17.9} & 9.61  & \textbf{15.1} \\
        Mamba2          &0.188&0.899&0.622&0.191&0.811& 29.1 & \textbf{9.11}  & 14.5 \\
        Scan $4\times$          &0.198&\textbf{0.916}&0.626&\textbf{0.173}&\textbf{0.841}& 21.6 & 14.0  & 9.65 \\ % +sa
        %Scan $4\times$          &TODO &TODO &TODO &TODO &TODO & \textbf{14.7} & 10.8  & 12.8 \\ % -sa
    \end{tabular}
    \caption{Average performance over the SC2 battle sequence and training cost.}
    \label{tab:sc2-perf}
\end{table}

\section{Discussion and Limitations}

Our benchmark sequences are performed over relatively short timespans. Accuracy over the \textit{Chasing-Targets} challenge saturates early into this cycle, however we note accuracy decay in some encoders (Fig. \ref{fig:accuarcy-time}). SC2 accuracy continues to increase until the 30 step cut-off. Micromanagement tasks may not elicit long range dependencies to the extent of short term ones, with players performing hundreds of actions per minute. There may be additional value in  the development and benchmarking of the models above on tasks with longer term  temporal dependencies and irregular and complex observation spaces.

\section{Related work}

\textbf{State Space Models} (SSMs) are ubiquitous in sequential modelling tasks. In control and dynamics modelling, they have been applied to learn locally linear latent dynamics models conditioned on exogenous inputs \citep{E2c,RCE,NVAE,PCC,DVBF,DeepKoopman}. More recently, these models have become popular with endogenous inputs. Here, SSMs parameterize a sequential process with endogenous input $\vu(t)$ to model the output process $\vy(t)$ with hidden state $\vx(t)$,
\begin{align}
    \dot{\vx}(t)=\mA\vx(t)+\mB\vu(t) \\
    \vy(t)=\mC\vx(t)+\mD\vu(t).
\end{align}
Parameters $(\mA,\mB,\dots)$ can be approximated in a discrete form $(\Bar{\mA},\Bar{\mB},\dots)$, appropriate for a sequence-to-sequence mapping, such that they can be learned with gradient descent. However, these matrices must be deliberately structured for effectiveness, usually guided by HiPPO theory \cite{gu2020hippo}. The key desirability of SSMs is re-parameterization as a convolution to enable parallel computation for training, 
\begin{align}
    \vy_t=\sum_{k=0}^t\bar{\mC}\bar{\mA}^k\bar{\mB}\vu_t + \bar{\mD}\vu_t\\
    \vy=\bar{\mK}*\vu+\bar{\mD}\vu,
\end{align}
where $\mK=(\bar{\mC}\Bar{\mA}^i\Bar{\mB})_{i=0\dots L-1}$ for sequence length $L$. Rather than learning the SSM matrices, \cite{fu2023simple} propose learning $\bar{\mK}$ directly with various regularization techniques and an efficient long convolution implementation to enable feasibility. \cite{gu2023mamba} introduce the notion of SSM matrices also being time dependent, similar to an LSTM, adding mechanisms for selectively propagating or forgetting state information dependent on the current token. This is also accompanied by a specialised implementation for performance. In follow up work, \cite{mamba2} decomposes the SSM into diagonal and low-rank blocks for computational efficiency. 

\textbf{Sub-quadratic Transformer Attention} has been proposed with various techniques. \cite{katharopoulos2020linattn} re-parameterise multi-head self-attention with a non-negative feature map (i.e. $\text{elu}(x)+1$) applied to $\mathbf{K},\mathbf{Q}$ separately to attain linear time and memory complexity, and enable incremental processing in scenarios with causal masking. RWKV \citep{peng2023rwkv} utilizes learned position biases, similar to \citep{zhai2021attention}, rather than queries from that position. \cite{sun2023retnet} follows a similar mechanism to our proposed method where a hidden state $\mathcal{S}$ is propagated with a decay factor $\gamma$ and updated with the current input, $\mathcal{S}_n=\gamma\mathcal{S}_{n-1}+\mathcal{X}_n$. However, our cyclic scanning and updating conditions the input on the accumulated history, before adding to the time-decaying latent state, as illustrated in Figure \ref{fig:enc-iscan}.

\textbf{Irregular Spatio-Temporal Tasks} can be found in many domains. Multi-object tracking is a popular task in computer vision, which aims to associate detected objects between frames. Recursive processing is popular in end-to-end learning of the multi-object tracking problem \citep{meinhardt2022trackformer,zhang2023motrv2,zhu2022looking}. This methodology is suitable for training short-term tracking algorithms, appropriate for many benchmarks \citep{dendorfer2020mot20,dave2020tao}. However, over a longer horizon, for example in long-term Re-identification tasks, recursive methods scale poorly, hence further innovation is required. Longer horizon methods \citep{qin2023motiontrack} are typically are not trained end-to-end like for this reason. Graph neural networks (GNNs) \citep{GNNs} are popular architectures for variable dimensional observations or irregularly structured data due to their permutation invariance. These models generally use zero-padding to handle irregular sized observations and rely on fixed graph structures for global context, although some approaches attempt to model the evolution of the graph over time \citep{evolvegcn, DynamicGCNs}. Unfortunately, these approaches lack the scalability of state space models. They are also difficult to frame in an incremental manner for inference, GNNs typically process the full sequence. For example, \cite{gao2020vectornet} construct a graph representation of a traffic scene with agent tracklets, lanes and crosswalks to produce a future trajectory estimation for the agents. GNNs such as these lack incremental inference, at each time-step the spatio-temporal graph is modified with the new observation, and re-processed in its entirety.

\section{Conclusion}

This work introduces a sequential modelling approach for irregular observation sources of varying dimensionality and cardinality. We investigate the use of various encoders to pack these observations into a fixed dimensional latent, and corresponding sequence modelling approaches. We propose a novel algorithm that alternates between cross attention and a weighted inclusive scan to accumulate context over time and show that this performs on par with existing sequence modelling approaches on two new challenging benchmarks, with comparably higher training and inference speeds and a lower parameter count. A key benefit of the proposed approach is that it can naturally be tweaked to allow a broad range of performance/compute tradeoffs (see Appendix \ref{sec:gym-detail-results} for more extensive experiments in this vein) as required for a given application.

\section{Reproducibility}

Model source code and experiment configuration files will be made public later. Chasing targets can be installed with pip \textit{chasing-targets-gym}, source code is at \url{https://github.com/5had3z/chasing-targets-gym}. StarCraft II uses the \textit{sc2-serializer} framework, \url{https://github.com/5had3z/sc2-serializer}, the tournaments dataset was used.

\bibliography{iclr2025_conference}
\bibliographystyle{iclr2025_conference}

\appendix
\section{Appendix}

\subsection{StarCraft II ROI Calculation} \label{sec:sc2-roi-calculation}

The objective of the ROI is to contain the main action happening on the screen, namely the units in combat. To achieve this, we gather the coordinates of all the units with target or motion commands over the sequence. K-Means clustering is performed on the positions and sorted by the number of members in the clusters. If the limits centroids of the K-means clusters can fit within the desired ROI size, then the mean of the centroids is used for the center of the ROI. If this condition is not fulfilled, the smallest cluster is removed. This is done iteratively until the condition is satisfied. This calculation can be performed live in the dataloader during training, or calculated once and saved a key associated with the sample to be re-indexed during training.

\subsection{StarCraft II Unit Positioning} \label{sec:sc2-position}

HL-Gauss \citep{pmlr-v80-imani18a} loss is used for categorical position encoding. Instead of a one-hot encoding label, a gaussian kernel is used as the categorical target with $\mu$ equal to the position and $\sigma$, which we set to $0.2$. The appropriate number of bins used for the prediction is correlated to the $\sigma$ chosen \citep{farebrother2024stop}, we use 20. We decode the categorical output of the model by integrating over the product of the soft-max output and the value each bin represents. We perform an ablation study with the large Scan $6\times$ encoder, shown in Table \ref{tab:sc2-pos-types}, to determine the best performing position learning schema. Based on these results, we use a relative cartesian coordinate system for unit position estimation, encoded with a categorical distribution.

\begin{table}[b]
    \centering
    \begin{tabular}{ccccccc}
\multirow{2}{*}{Rep.}&\multirow{2}{*}{Frame}&\multirow{2}{*}{Encoding}&\multicolumn{4}{c}{MSE}\\
                    &                       &                         &  0     & 5     & 15    & 25    \\
    \hline
    % Polar   & Rel. & Scalar &  & & & \\ Bad data
    Cart.   & Rel. & Scalar & 0.300 & 0.258 & 0.253 & 0.251 \\
    Cart.   & Glbl.& Scalar & 0.299 & 0.259 & 0.254 & 0.241 \\
    Polar   & Rel. & Cat.   & 0.258 & 0.221 & 0.216 & 0.216 \\
    Cart.   & Rel. & Cat.   & \textbf{0.248} & \textbf{0.211} & \textbf{0.207} & \textbf{0.206} \\
    Cart.   & Glbl.& Cat.   & 0.254 & 0.220 & 0.218 & 0.219 \\
    \end{tabular}
    \caption{Position representation ablation performed with Scan $6\times$ encoder. Here Cart. is cartesian, Rel. is relative position from the unit, Glbl. is global position in the ROI and Cat. is categorical position encoding.}
    \label{tab:sc2-pos-types}
\end{table}

In this section we discuss how we handle wrapping $\theta$ when using a polar coordinate system with a categorical representation. When creating the categorical target, we double the template range to $[-4\pi,4\pi]$. Once the kernel is projected onto the template, we add the reflection of the additional range to the opposite side in the normal range, i.e. we reflect and add the bins from $[-4\pi,-2\pi]$ to $[0,2\pi]$, and repeat for the the opposite tail. To handle angle wrapping during decoding, we find the argmax category and rotate the categories so the argmax is centered. With this new categorical range centered on the argmax, the values each bin represents is recalculated to the new range. The standard decoding algorithm is used to yield the final result.

\subsection{Recurrent Model Comparison} \label{sec:rnn-compare}

We evaluated all the available recurrent neural network modules available in PyTorch, and briefly compare a learned or zero initialization for the LSTM. The standard RNN performs poorly whilst the LSTM and GRU are similar. Learned or zero initialization for the LSTM also has minimal impact on the performance of the model on the \textit{Chasing-Targets} benchmark, with the learned initialisation having a slightly greater peak, but zeros climbing to higher accuracy faster. Each of these models use the X-Attn encoder with piece-wise encoding algorithm.

\begin{figure}
    \centering
    \begin{tikzpicture}
\begin{axis}[
    height=3.5cm,
    width=10cm,
    ylabel={Top1},
    xmin=0, xmax=40,
    ymin=0.5, ymax=0.75,
    xtick={5,10,15,20,25,30,35},
    ytick={0.55,0.6,0.65,0.7},
    legend columns=1,
    legend pos=outer north east,
    legend style={
        font=\footnotesize,
        legend cell align=left,
    },
    ymajorgrids=true,
    grid style=dashed,
    every axis plot/.append style={ultra thick},
    ]
\addplot[color=lightred] table [x index=0, y index=1, col sep=comma]{data/rnn-val-gym-Top1.csv};
\addlegendentry{LSTM L}
\addplot[color=darkorange] table [x index=0, y index=2, col sep=comma]{data/rnn-val-gym-Top1.csv};
\addlegendentry{LSTM Z}
\addplot[color=mediumblue] table [x index=0, y index=3, col sep=comma]{data/rnn-val-gym-Top1.csv};
\addlegendentry{RNN L}
\addplot[color=darkblue] table [x index=0, y index=4, col sep=comma]{data/rnn-val-gym-Top1.csv};
\addlegendentry{GRU L}
\end{axis}
\end{tikzpicture}
    \caption{Comparison between recurrent models and hidden state initialization where Z=zeros and L=learned.}
    \label{fig:rnn-compare}
\end{figure}
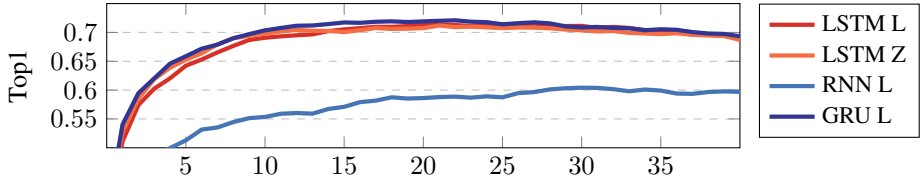

\subsection{Chasing Targets Detailed Results} \label{sec:gym-detail-results}

Table \ref{tab:gym-main} provides a larger set of results including various encoder (BERT, X-Attn) and encoding algorithm (Fused, Piece-wise) variants for the S.T, Recurrent and Mamba2 encoders. We also include slight variations of the Scan encoder with removed self-attention layers to trade accuracy and efficiency (indicated by $^1$) or replace cross-attention with the observation in the repeat layers, with self attention blocks (indicated by $^2$).

\begin{table}[]
    \centering
    \begin{tabular}{ccccccccc}
\multirow{2}{*}{Encoder}& \multicolumn{4}{c}{Top1 Accuracy} & \# M  & Mem. & Train  & Infer. \\
                    & 0     & 5     & 10    & 40     & Param.& GB   & it/sec & it/sec \\
\hline
    RNN-L,Fused,BERT & 0.359 & 0.529 & 0.592 & 0.633 & 0.76 & 1.82  & \textbf{60.7} & \textbf{89.4} \\ % d8dffc193fb4c0960d5b7ea3801e51db 
    GRU-L,Fuse,BERT & 0.372 & 0.652 & 0.706 & 0.711 & 1.29 & 1.88  & 58.8 & 86.6  \\ % ffc53cdda8ae4987feb440f06b114d10
    LSTM-L,Fuse,BERT& 0.375 & 0.646 & 0.703 & 0.703 & 1.56 & 1.89  & 56.4 & 81.4  \\ % 0ac7f2a9ce5fcd33881183478cb37fce
    S.T,PW,BERT      & 0.377 & 0.688 & 0.739 & \textbf{0.719} & 2.10 & 2.81  & 35.3 & 55.2  \\ % 171606a43ebc623b5f800ff01b38e57b
    S.T,PW,X-Attn    & 0.374 & 0.692 & \textbf{0.745} & 0.716 & 2.10 & 1.86  & 42.7 & 52.1  \\ % 6548c36ec4a34985473458d8fd2271b4
    Mamba2,PW,X-Attn & 0.381 & 0.673 & 0.724 & 0.703 & 2.79  & 1.04  & 30.9  & 31.6  \\ % f8df2c61d04c5f4716f7a40e8b882071
    Mamba2,Fuse,BERT & 0.386 & 0.667 & 0.719 & 0.713 & 2.79  & 1.95  & 35.4  & 61.0  \\ % 118a9454d17b581571afb260d0214c3d
    % Recursive       & 0.336 & 0.543 & 0.615 & 0.646 & 1.78  & 2.86  & 2.28 & 2.20  \\ % ? 63ea5c801a62d6ff528af7839af2950b
    Scan $1\times$   & 0.358 & 0.514 & 0.557 & 0.600 & 0.41 & 0.74  & 77.7 & 92.5  \\ % fb4ba5c665f959c5197676270a321b7a
    Scan $2\times$   & 0.334 & 0.658 & 0.688 & 0.712 & 0.65 & 1.13  & 49.7 & 54.8  \\ % a8cf5bc3bc6162a50f2382b8a7f6dd4e
    Scan $4\times^1$ & 0.399 & 0.629 & 0.668 & 0.677 & 1.31 & 1.82 & 30.4 & 33.3 \\ % a6178f190e01629aef1bf0c6d7c08e57 (no resample)
    Scan $4\times^2$ & 0.368 & 0.661 & 0.696 & 0.705 & 0.91 & 1.47 & 39.7 & 41.3 \\ % 4e6242c12ce09b2a0651f9a11e687c03 (-sa)
    Scan $4\times$   & 0.364 & 0.680 & 0.717 & 0.704 & 1.31 & 1.95 & 29.2 & 30.6  \\ % f4309fa3d3d93e702ea6848241237817
    Scan $6\times$   & \textbf{0.403} & \textbf{0.696} & 0.723 & 0.703 & 1.91 & 2.79  & 20.7 & 21.7  \\ % c78ee4774a0a096d1d3c8df023897543
    \end{tabular}
    \caption{Algorithm performance, training and evaluation cost of tested algorithms on \textit{Chasing-Targets}. $^1$ A self-attention layer is skipped in the encoder block. $^2$ The input is only sampled once.}
    \label{tab:gym-main}
\end{table}

\end{document}